\documentclass[11pt,oneside,a4paper]{amsart}

\usepackage[T1]{fontenc}
\usepackage{color}
\usepackage[utf8]{inputenc} 
\usepackage[pdftex]{thumbpdf}

\usepackage{amsthm}
\usepackage{amsmath}
\usepackage{amsrefs}
\usepackage{paralist}
\usepackage{mathtools}
\usepackage{colonequals}
\usepackage{breqn}
\usepackage{amssymb}

\usepackage{graphicx}
\usepackage{caption}
\usepackage{verbatim}
\usepackage{setspace}
\usepackage{lipsum}
\usepackage{algorithm}
\usepackage{algorithmicx}
\usepackage{algpseudocode}
\usepackage{comment}

\theoremstyle{plain}
\newtheorem{theorem}{Theorem}[section]

\newtheorem{proposition}[theorem]{Proposition}

\newtheorem*{theorem*}{Theorem}
\newtheorem*{lemma*}{Lemma}
\newtheorem*{prop*}{Proposition}
\newtheorem*{cor*}{Corollary}
\newtheorem*{conj*}{Conjecture}

\theoremstyle{definition}
\newtheorem*{definition*}{Definition}
\newtheorem{definition}[theorem]{Definition}

\theoremstyle{remark}
\newtheorem*{rem*}{Remark}
\newtheorem{rem}[theorem]{Remark}
\newtheorem*{example*}{Example}

\numberwithin{theorem}{section}
\numberwithin{equation}{section} 
\numberwithin{figure}{section} 

\newcommand{\R}{\mathbb{R}}

\newcommand{\ceq}{\colonequals}

\title{A Topological Approach to Spectral Clustering}
\author{Antonio Rieser}
\date{}
\address{CONACYT-CIMAT, Centro de Investigación en Matemáticas, Calle Jalisco
S/N, Colonia Valenciana, C.P 36023 Guanajuato, Guanajuato, Mexico}
\email{antonio.rieser@cimat.mx}
\thanks{This research was supported in part by the grant TOPOSYS FP7-ICT-318493-STREP, Cátedras CONACYT 1076, the US Office of Naval Research Global, and the Southern Office of Aerospace Research and Development of the US Air Force Office of Scientific Research}

\begin{document}

\begin{abstract} 
We propose two related unsupervised clustering algorithms which, for input, take data
assumed to be sampled from a uniform distribution supported on a metric space
$X$, and output a clustering of the data based on the selection of a
topological model for the connected components of $X$. Both algorithms work by selecting a graph on the
samples from a natural one-parameter family of graphs, using a
geometric criterion in the first case and an information theoretic
criterion in the second. The estimated connected components of $X$ are
identified with the kernel of the associated graph Laplacian, which allows the
algorithm to work without requiring the number of expected clusters or other
auxiliary data as input.
\end{abstract}

\maketitle

\section{Introduction}

The analysis of complex, high-dimensional data is one of the major research
challenges in contemporary computer science and statistics. In recent years,
geometric and topological approaches to data analysis have been shown to yield
important insights into the structure of complex data sets. The common point
of departure in these methods is the assumption that data in high-dimensional
spaces is often concentrated around a low-dimensional manifold or other
topological space. 

Geometric techniques, in particular, have proven to be particularly successful.
These have largely concentrated on approximating the
local geometry of the data as a step towards non-linear dimension reduction.
Once an embedding of the data in a lower-dimensional space has been found, standard
statistical techniques are then used to analyze the data in the
lower-dimension. Methods in this class include ISOMAP
\cite{Tenenbaum_de_Sliva_Langford_2000}, Locally Linear Embedding
\cite{Roweis_Saul_2000}, Hessian Eigenmaps \cite{Donoho_Grimes_2003}, Laplacian
Eigenmaps \cite{Belkin_Niyogi_2003}, and Diffusion Maps
\cite{Coifman_Lafon_2006}.  Most of these techniques build a weighted graph to
approximate the Laplace-Beltrami operator on a manifold, or else a related
Markov chain on a graph, and then use the eigenvalues and
eigenvectors of the resulting operator to reduce the dimension of the data,
often, in practice, followed by the application of a $k$-means clustering
algorithm to perform the clustering of the data \cite{Shen_et_al_2014},
\cite{Sanchez-Garcia_et_al_2014}. We encounter a more topologically-oriented approach in
\cite{Chazal_Guibas_2013}, in which persistent homology
\cites{Carlsson_Zomorodian_2005, Ghrist_2008, Carlsson_2009} is used to help a
statistician identify high-density regions of a distribution function.

In this article, we give an algorithm that directly uses the topological
information in the graph Laplacian to produce a clustering of the
data, eliminating the need for separate dimension reduction and clustering
steps. This process, furthermore, illustrates the utility of considering the data clustering
problem from a topological, instead of purely analytic, perpective, i.e. to
consider clustering as a problem of estimating the number of
connected components of the support of an idealized underlying distribution. While the
topological aspect of the clustering problem has been generally acknowledged in the
topological data analysis community for some time, this is, to the best of our knowledge, the
first completely data-driven clustering algorithm that explicitly exploits this
point of view. Additionally, the algorithm produces both the number of clusters and
the clusters themselves with no additional information required, unlike popular algorithms such as $k$-means
clustering or $k$-nearest-neighbor clustering, in which the number of clusters
or other additional input is required. Finally, we
propose what we believe to be the first geometric and information-theoretic
model selection criteria for choosing a Laplacian (or, more precisely, the
associated Markov process) from a family of
candidates built from the geometric properties of the sample space
alone, instead of from observations of the Markov process.

\section{Topology and Clustering}
\label{sec:TopC}

In all that follows, we will assume that there is an ideal unknown probability distribution $P(X)$ supported 
on a disconnected metric space $X = \sqcup_{i=1}^k X_i$, where $X$ is embedded as a (possibly much lower dimensional) 
subset of an ambient space $Y$. We will further suppose that the distribution $P(X)$ may be corrupted by noise, 
and that the coordinates of the sample points are given by sampling from the combined noisy distribution. 
We will consider a clustering to be correct if it accurately recovers the number of
connected components of $X$ and it assigns to each sample point to the nearest connected component. 

When $X$ has the homotopy type of a finite CW-complex, the number of connected
components of $X$ is given by the dimension of the
$0$-th real cohomology group $H^0(X;\R)$. Furthermore, if $X$ is a manifold, this is equal
to the dimension of the kernel of the Laplace-Beltrami operator $\Delta_X$
through a small amount of Hodge theory \cite{Jost_2011}. We recall that the Laplace-Beltrami operator 
$\Delta:L^2(X) \to L^2(X)$ may be defined by $\Delta = -d^*d$, where $d$ is the exterior derivative,
extended continuously to $L^2(X)$, and $d^*$ is its adjoint in $L^2(X)$. It
follows, too, that the functions in the kernel are constant on each connected
component of $X$ (Lemma 3.3.5 in \cite{Jost_2011}).

The following proposition now follows easily.

\begin{proposition}
\label{prop:Image F}
Let $(X,g)$ be a smooth Riemannian manifold, and let $k$ denote the
number of connected components of $M$. Let $\{\psi_i\}_{i=1}^{k}$ be a basis of the
kernel of the Laplace-Beltrami operator $\Delta_X$ of $X$, and define the map
$\Psi:X \to \R^{k}$ by
\[\Psi(x) \ceq (\psi_1(x),\dots,\psi_{k}(x)) \in \R^{k}.\]
Then the image of $\Psi$ consists of exactly $k$ points in $\R^{k}$, and the image of each connected component of $X$ is a single point.
\end{proposition} 
\begin{proof} First, we know from \cite{Jost_2011}, Section 3.3, that each basis
	function $\psi_i$ is constant on each connected component of $X$, and it follows
	that each connected component is sent to a single point. It only
	remains to show that no two connected components are sent to the same
	point. Consider the matrix $A$ defined by $a_{ij} = \psi_i(X_j)$. Since
	the $\psi_i$ are linearly independent, this is a $k \times k$ matrix of
	full rank. Suppose now that there are two connected components, $X_1$
	and $X_2$, whose image under $F$ is the same point $x \in
	\R^{k}$. Then two rows of $A$ are the same, and the $\text{rank }A <
	k$, a contradiction. Therefore, all of the connected components of $X$
	are sent to different points in $\R^{k}$.
\end{proof}

\begin{rem}
	We note, too, that Proposition \ref{prop:Image F} also hold for graphs and the graph Laplacian instead of manifolds, with a nearly identical proof.
\end{rem}

\section{The Graph Laplacian and the Graph Heat Semigroup}

Given a noisy sample from a disconnected metric space $X$, motivated by the
above discussion, our primary task in the clustering problem will be
to compute an empirical Laplacian $\hat{\Delta}$ and the corresponding empirical function $\hat{\Psi}$ so that
$\hat{\Psi}$ is constant on the points sampled from a given connected component
of $X$. We begin by recalling the construction of the graph Laplacian, the
associated heat semigroup, and several fundamental results.

Let $G=(V,E,w)$ be a weighted graph, where the \emph{weight function} $w:E \to
[0,\infty)$ gives the
weights of every edge. We define the matrix $L_{G}$ to be
\begin{equation}
	(L_{G})_{(i,j)} =
\begin{cases}
w(x_i,x_j) & i\neq j,\\
-\sum_{v_j \in V} w(x_i,x_j) &  i = j, \\
\end{cases}
\end{equation}
where $x_i,x_j \in V$. We likewise define the corresponding heat operators $e^{-tL_G}$ by
\begin{equation*}
e^{-tL_G} = \sum_{k=0}^\infty \frac{(-tL_{G})^k}{k!}.
\end{equation*}
for $t \in [0,\infty)$. Note that $e^{-tL_G}e^{-sL_G} = e^{-(t+s)L_G}$, so the
set of matrices $\{e^{-tL_G}\}_{t \in [0,\infty)}$ forms a semigroup under matrix
multiplication, which we call the \emph{graph heat semigroup}, or the
\emph{heat semigroup of G}. The following result is immediate from the definitions.

\begin{proposition}
	\label{prop:Graph eigvals and eigvecs}
	The vector $v\in \R^{|V|}$ is an eigenvector of $L_G$ with eigenvalue
	$\lambda$ iff $v$ is an eigenvector of $e^{-tL_G}$ with eigenvalue
	$e^{-t\lambda}$.
\end{proposition}

As in the manifold case, we have the following result.

\begin{theorem}[\cite{Chung_1997}, Lemma 1.7.iv, and page 3, equation 1.1]
	\label{thm:Kernel dimension and connected components}
The number of connected components of $G$ is equal to the dimension of the
kernel of $L_G$, and the vectors $v \in \ker L_G$ are constant on each
connected component of $G$, i.e. if the vertices $x_i, x_j \in V$ are in the same connected
component of $G$, then for the corresponding coordinates $v$ we have $v_i =
v_j$.
\end{theorem}
\begin{rem*} Combining Proposition \ref{prop:Graph eigvals and eigvecs} with
	Theorem \ref{thm:Kernel dimension and connected components}, we see
	that the dimension of the $1$-eigenspace of $e^{-tL_G}$ is also equal
	to the number of connected components of $G$, for any $t \in
	(0,\infty)$. 
\end{rem*} 

\section{Graph Models and Model Selection for the Graph Heat Semigroup}
\label{sec:Graph models and model selection}

We now describe how to combine the above ideas with several additional
observations to perform unsupervised distance-based clustering
on a data set $Z$. 

\begin{definition}
\label{def:Family of graphs}
Let $Z$ be a collection of points in a metric space $Y$. For
each $r \geq 0$, we define a graph $G_r = (V_r,E_r)$ by
\begin{align*}
	V_r &= Z\\
	E_r &= \{(x_i,x_j) \in Z \times Z \mid d_Y(x_i,x_j) \leq r\}.
\end{align*}
\end{definition}
That is, the vertices for each graph $G_r$ are the points in the data
set $Z$, and two vertices in $G_r$ are joined by an edge iff the distance between them is
at most $r$.

Once we have a collection of graph models $\{G_r\}_{r\geq 0}$ for the data set
$Z$, we compute their corresponding graph
Laplacians $L_{G_r}$ (which we abbreviate to $L_r$), and the heat semigroups $\{e^{-tL_r}\}_{t \in
[0,\infty)}$. We would now like choose a value $\hat{r}$ from among the $r\geq
0$ so that
$\{e^{-tL_{\hat{r}}}\}_{t \in [0,\infty)}$ best represents the heat semigroup $\{e^{-t\Delta_X}\}_{t \in
[0,\infty)}$, where $X$ is the support of the
distribution of the process which generated the points. To solve
the clustering problem, it is sufficient to choose a $\hat{r}$ so that the dimensions of the kernels of $L_{\hat{r}}$ and
$\Delta_X$ are equal, i.e. so that the graph $G_{\hat{r}}$ has the same number
of connected components as $X$, and then
assign each vertex to the correct connected component. Note that, if we
choose $r$ too small, then there will be too many small connected components,
but if we choose $r$ too large, then there will be too few large ones.

We give two techniques for solving this problem, the first based on a geometric
criterion, and the second based on an information-theoretic one. We compare their
performance in Section \ref{sec:Examples}.

\subsection{Model Selection by Average Relative Neighborhood Volume}
\label{subsec:Average Relative Neighborhood Volume}

Let $M$ be a disconnected closed manifold (i.e. without boundary), let $x\in M$
be a point in $M$, and denote by $M_x \subset M$ the
connected component of $M$ containing $x$. Let $|A|$ denote the volume of
$A$ for any $A\subset M$ using the volume form on $M$. Our first technique for choosing $r$ is based on the observation that, on any disconnected manifold $M$, where each
connected component is of dimension $> 0$, for any
point $x \in M$, the function 
\[
	R(x) \coloneqq \lim_{U \in \mathcal{N}(x)} \frac{|U|}{|M_x|} = 0
\]
where the limit is taken over the net defined by the partially ordered set $\mathcal{N}(x)$ of
neighborhoods of $x$. That is, for any $x\in M$, the ratio of the volume of a neighborhood of $x$ to the volume
of the connected component containing $x$ may be made arbitrarily small. It follows that
\[
	R_M \coloneqq \frac{1}{|M|}\int_M R(x) \, dx = 0
\]
as well.

Now consider the family of graphs $G_r$ built on a fixed, finite sample
of points $Z$ taken from
a uniform distribution on a metric space $X$ as in Definition \ref{def:Family
	of graphs}. We say that a subset $U\subset V_r = Z$ is a
neighborhood of a vertex $x \in Z$ iff $U$ contains $x$ and all vertices
adjacent to $x$, and let the volume of a subset of $V_r$ equal its cardinality.
Note that the volume of any
neighborhoods $U$ of a point $x$ is necessarily bounded below by $1$, and furthermore,
if $M_x = \{x\}$, then $R(x) = 1$ as well. It
follows that, for $r$ sufficiently small, $R(x) = 1$ for any $x$ (since every
vertex is its own connected component in $G_r$), and therefore
$R_{G_r} = 1$ as well. On the other extreme, if $r$ is sufficiently large, then
$G_r$ will
have only one component, and the smallest neighborhood of any vertex $v$ is
$V$. In this case, we also have $R_{G_r} = 1$. It is not, however, difficult to find
graphs $G$ with $R_G < 1$. For example, consider the circular graph $G_C =
(V_C,E_C)$ with 
\begin{align*}
	V &= \{0, \dots, n-1\}\\
	E &= \{(i \text{ mod } n,(i+1) \text{ mod } n) \mid i \in V\}.
\end{align*}
We therefore have that $R_{G_C} = 3/n$, and therefore $R_{G_C} < 1$ for any $n
> 3$.

We define
\begin{align}
	\hat{r} = \text{argmin }R_{G_r},
\end{align}
i.e. $\hat{r}$ is the value of $r$ such that the graph $G_r$ has minimal $R_{G_r}$. In
the case of a tie, we take the smallest $r$. 

In this method, $G_{\hat{r}}$ will be our choice of graph model for the data set $Z$, with
Laplacian $L_{\hat{r}}$ and heat semigroup $\{e^{-tL_{\hat{r}}}\}_{t \in
[0,\infty)}$.

\subsection{Model Selection by Average Relative Entropy}
\label{subsec:Average Relative Entropy}

We now give an information-theoretic criterion for choosing $r$, which we
motivate with the following discussion. As in the case of the Average Relative
Neighborhood Volume Criterion in Section \ref{subsec:Average Relative Neighborhood
Volume}, we wish to choose a graph $G_r$ which is sufficiently locally
connected to recover the connected components of $X$, but no more. 
Not that, if $\phi^y_0$ is a delta distribution centered at the point
$x\in M$ in a manifold $M$, then the resulting steady state $\phi^{y}_{*}$ of the heat flow
with initial condition $\phi^y_0$ is constant on the connected component of $M$ containing
$x$, and in particular, $\phi^{y}_{*}(x) = \frac{1}{|M|}$ for all $x \in M$. Using
the fact that $\phi_t$ is a probability distribution for each $t$, we also note
that
\begin{align*}
	H(\phi^{y}_{*}) - H(\phi^y_t) &= \int_M \phi^y_t(x) \ln(\phi^y_t(x)) \,dx - \int_M
	\phi^{y}_{*}(x)
	\ln(\phi^{y}_{*}(x))  \, dx\\
	&= \int_M \phi^y_t(x) \ln(\phi^y_t(x)) \, dx - |M| \phi^{y}_{*}(x) \ln(\phi^{y}_{*}(x))\\
	&= \int_M \phi^y_t(x) \ln(\phi^y_t(x)) \, dx - \ln(\phi^{y}_{*}(x))\\
	&= \int_M \phi^y_t(x) \ln(\phi^y_t(x)) \, dx - \ln(\phi^{y}_{*}(x)) \int_M
	\phi^y_t(x) \, dx\\
	&= \int_M \phi^y_t(x) \ln(\phi^y_t(x)) \, dx - \int_M \phi^y_t(x)
	\ln(\phi^{y}_{*}(x))\, dx\\
	&= d_{KL}(\phi^y_t,\phi^{y}_{*}),
\end{align*}
where $H(\phi) = \int_M \phi(x) \ln (\phi(x)) \, dx$ is the entropy of the
distribution $\phi$, $d_{KL}$ is the Kullback-Leibler divergence, or relative entropy, between $\phi_1$ and $\phi^{y}_{*}$.

For a distribution $\phi^y_t$ which is concentrated around a point $y$ on a
large connected component $M_y$, $d_{KL}(\phi_t,\phi^{y}_{*})$ will be large,
and furthermore, if this is true for any $y\in M$, it will also true for the average
\begin{equation*}
	H_{M,t} \coloneqq \frac{1}{|M|} \int_M d_{KL}(\phi_t^y,\phi^{y}_{*}) \, dy.
\end{equation*}

Now, let $\{G_r\}_{r\geq 0}$ be the family of graphs from Definition \ref{def:Family
of graphs}, $\{L_r\}_{r \geq 0}$ be, and we consider the collection of graph heat semigroups $\{e^{-tL_r}\}_{t \in
[0,\infty),r\geq 0}$, where the vertices of the $G_r$
are sampled from a uniform distribution on a manifold $X$ in $Y$ with noise. The empirical
initial distributions $\hat{\phi}^i_{r,0}$ centered at a point $i\in V$, corresponding to the delta distribution
in the manifold case, are given by the standard basis vectors
$e_i$, $i \in \{1,\dots,n\}$, where $n = |V_r|$, the number of vertices in the
graph. The solution $\hat{\phi}^i_{r,t}$ at time $t$ of the empirical heat flow with
initial condition $\hat{\phi}^i_{r,0} = e_i$ is therefore given by the $i$-th column of
$e^{-tL_r}$. The empirical steady state $\hat{\phi}^{i}_{r,*}$, naturally, is the
$i$-th column of $\lim_{t\to \infty} e^{-tL_G}$.

Define
\begin{equation*}
	H_{r,t} \coloneqq \frac{1}{n}\sum_{i=1}^n
	d_{KL}(\hat{\phi}^{i}_{r,t},\hat{\phi}^{i}_{r,*}).
\end{equation*}
Motivated by the discussion in the paragraphs above, we
will choose our preferred scale $\hat{r}$ to be the one which maximizes $H_{r,1}$. This
is, we choose $\hat{r}$ to be the value of $r$ at which the emprical
average relative entropy at time $t=1$
takes its maximum, i.e.
\begin{equation}
	\hat{r} \coloneqq \text{argmax } \hat{H}_{r,1}.
\end{equation}

When there are many small clusters $\bar{H}_{r,1}$ will be small, since the average size of the
support of $\hat{\phi}^{i}_{\hat{r},1}$ is a large portion of each connected
component. This will also be the case when $r$ is large. We therefore see that a kind of "bias-variance" tradeoff is built
into the geometry of local neighborhoods vs. connected components, and that
this is what powers both
methods.

We conclude this section with the remark that, from the properties of the heat equation on $\R^n$, the fundamental
solutions have maximal entropy among all distributions which satisfy certain
mean and variance constraints. One might expect, based one this, that we should
seek to minimize $H(\phi_{r,*})-H(\phi_{r,t})$, but, in fact, this is the opposite of
the effective approach, the relevant constraints on the family
$\{\phi_{r,t}\}_{r\geq 0}$ being different.

\section{Cluster Identification}

The results in Section \ref{sec:TopC} tell us that the points in each connected
component of a graph $G$ should be sent to exactly the same point in
$\R^{k}$, where $k$ is the number of connected components of $G$, by the map
$\Psi$. In practice, however, there are sometimes
small amounts of numerical error in the algorithms for computing eigenvalues and eigenvectors, and this must be accounted for when constructing the final clustering. We do this with a modified version of Gaussian elimination on the matrix formed by the eigenvectors, which we now describe.

First, note that the $j$-th entry in the eigenvector $f_i$ is the value of the eigenfunction $f_i$ evaluated on the point $z_j$. Let $\Psi$ be the matrix defined by
\begin{equation*}
	(\Psi)_{(i,j)}  = (\psi_i)_j = \psi_i(z_j),
\end{equation*} 

We give a modified Gaussian elimination algorithm in Algorithm \ref{alg:Gauss}.
For what follows, let $n$ denote the number of points in our sample, and let $k$
the number of connected components of the graph $G_{\hat{r}}$.
\begin{algorithm}
\caption{Modified Gaussian elimination on $\Psi$}
\label{alg:Gauss}
\begin{algorithmic}[1]

\For{$i = 1$ to $k$}
\State Reorder columns $i$ through $n$ of $\Psi$ so that $\vert \Psi_{(i,i)} \vert$ is the maximum of $\vert \Psi_{(i,j)} \vert$ in row $i$.
\State Divide row $i$ by $\Psi_{(i,i)}$
\State Using elementary row operations, make $\Psi_{(k,i)} = 0$ for $k\neq i$.
\EndFor
\State Redefine $\psi_i \ceq \Psi_{i,*}$, and (abusing notation) using the new
$\psi_i$, redefine the
map $\Psi(z_m)
\ceq (\psi_1(z_m),\dots,\psi_{k}(z_m))$
\end{algorithmic}
\end{algorithm}

Note that the algorithm, if there was no estimation error, would send each
point in the sample to one of the vectors $e_i$ in the standard basis of $\R^{k}$. Now, however, even given some numerical error, we are able to cluster the sample points according to how close $\Psi(x)$ are to each of the vectors $e_i$. 

\section{Algorithm and Experiments}
\label{sec:Examples}
We now give the complete algorithm and the results of some numerical
experiements. We denote the set of points by $Z$.

\begin{algorithm}[H]
\caption{Clustering algorithm}
\begin{algorithmic}[1]
\State For each $r < Diam(Z)$, compute $G_r$, $L_{r}$, $e^{-L_{r}}$ and
estimate $\lim_{t\to \infty} e^{-t L_{r}}$ by $e^{-t^* L_{r}}$ for some $t^*$ large (we use
$t^*=1000)$.
\State Using one of the methods in Sections \ref{subsec:Average Relative
Neighborhood Volume} or
\ref{subsec:Average Relative Entropy}, compute $\hat{r}$
\State Compute the kernel of $L_{\hat{r}}$, $\psi_i$, $i \in 1\dots k$
\State Using Algorithm \ref{alg:Gauss}, create the map $\Psi: z_m \mapsto
\Psi(z_m)
= ((\psi_1(z_m),\dots,(\psi_{k}(z_m)) \in \R^k$
\State Compute the distances $d_i(z_m) = \Vert \Psi(z_m) - e_i \Vert$ for each
point $z_m$ in the sample.
\State Assign the vertex $m$ to the $i$-th cluster if $d_i(z_m) < d_j(z_m)$ for all $j \neq i$.
\end{algorithmic}
\end{algorithm}

The following figures summarize the output of this algorithm on a data set of
500 points sampled with a small amount of Guassian noise from three interlinked circles
embedded in $\R^3$. The horizontal circle has radius $1$ and center $(0,0,0)$, and the other two
have radii $0.5$ and $0.4$ and centers $(0,-1,0)$ and $(0,1,0)$, respectively.
We ran several trials with different levels of Gaussian noise, with standard deviations $0.02$ (low noise), $0.03$ (medium
noise), and $0.045$ (high noise). The
value of $\hat{r}$ given by both methods was identical for the low noise
experiment, and the algorithm correctly assigned all of the points to their
respective circles. In the medium noise experiment, the Average Relative
Entropy Method
continues to classify the circles correctly, but the Average Neighborhood
Volume Method simply puts all of the circles into a single cluster. For the
high noise experiment, we see that the Average Relative Entropy Method also
starts to break down, but nonetheless identifies subclusters of the circles,
from which a user would be able to reconstruct the original circles. In
contrast, the Average Local Volume Ratio groups all of the points into a single
cluster. We
therefore see that, while both methods work in ideal conditions, the Average Relative Entropy method is
far more robust to noise. Interestingly, too, both the relative entropy curves
and the local volume ratio curves
exhibit local maxima and minima, respectively, where the smaller circles join the same group as the
central circle, behavior reminiscent of the "barcodes" in topological data
analysis. We include the figures below.

\subsection{Low Noise Experiment ($\sigma = 0.02$)}

\begin{figure}[H]
\includegraphics[width=\textwidth]{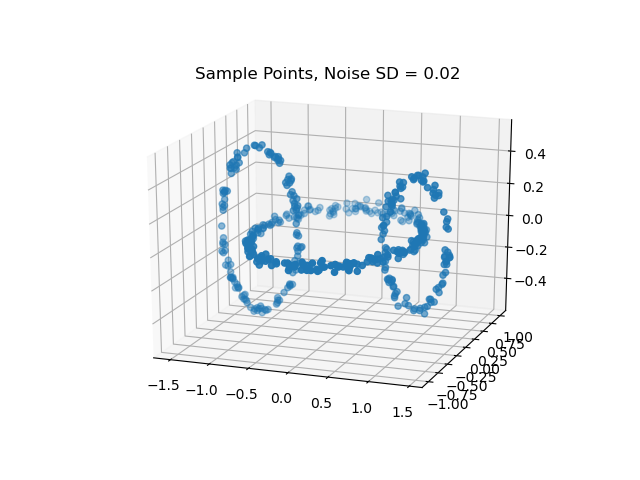}
\caption{500 points sampled from three circles with Gaussian noise ($\sigma =
0.02$)}
\label{fig:Variance20Points}
\end{figure}

\begin{figure}[H]
\includegraphics[width=\textwidth]{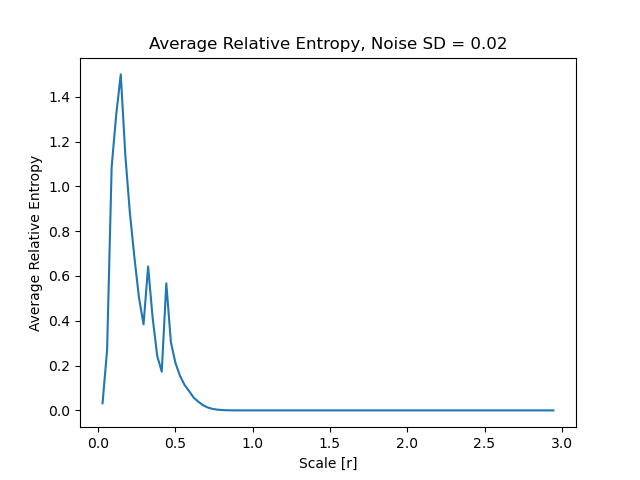}
\caption{Average Relative Entropy vs. Scale, Low Noise Experiment. Note that
local maxima appear where the smaller circles join to a larger cluster.}
\label{fig:Variance20RelEntPlot}
\end{figure}

\begin{figure}[H]
\includegraphics[width=\textwidth]{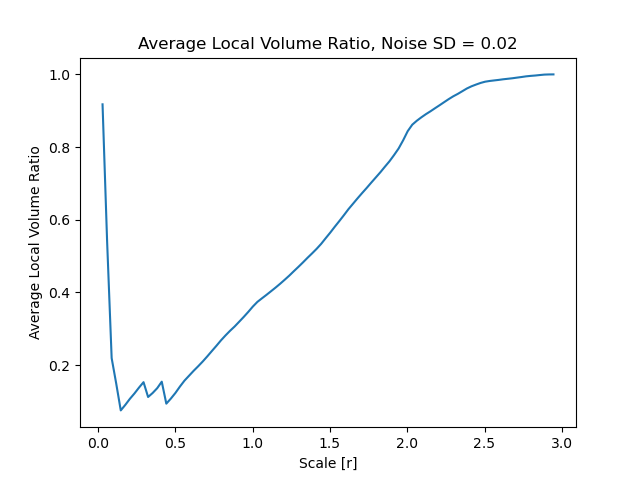}
\caption{Average Local Volume Ratio vs. Scale. Note that local minima appear
where the smaller circles join with the larger circle.}
\label{fig:Variance20ALVRPlot}
\end{figure}

\begin{figure}[H]
\includegraphics[width=\textwidth]{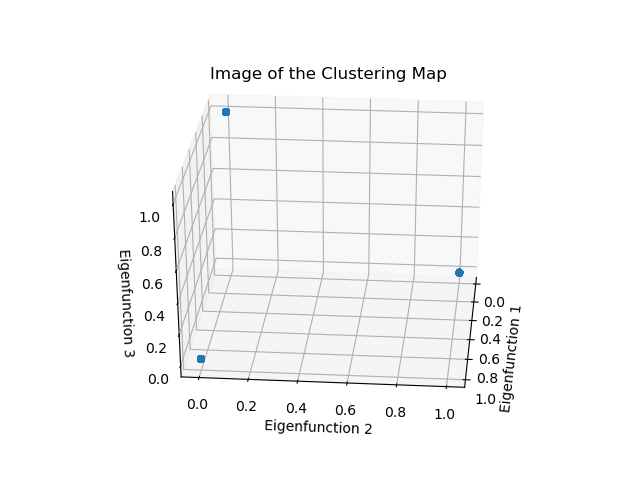}
\caption{The image of $\Psi$, centered around the points $(1,0,0), (0,1,0),$ and $(0,0,1)$}
\label{fig:Variance20PsiPlot} 
\end{figure}

\begin{figure}[H]
\includegraphics[width=\textwidth]{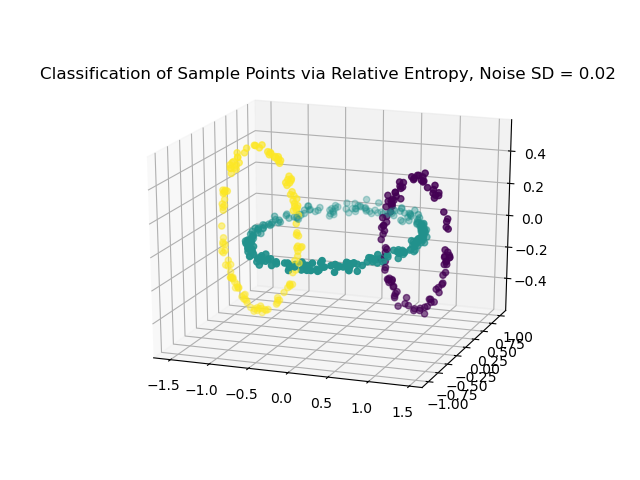}
\caption{The classification of the points using the Average Relative Entropy
method, illustrated by color. The classification produced by the Average Local
Volume Method was identical in this experiment.}
\label{fig:Variance20Clusters}
\end{figure}

In this experiment, both methods successfully recovered the circles. Figure
\ref{fig:Variance20Points} illustrates the data set, Figure
\ref{fig:Variance20RelEntPlot} shows the Average Relative Entropy at each
scale, Figure \ref{fig:Variance20ALVRPlot} gives the plot of the Average Local
Volume Ratio at each scale, Figure \ref{fig:Variance20PsiPlot} shows the image of the
map $\Psi$, and, finally, Figure \ref{fig:Variance20Clusters} shows the
classification of the points. Note that in Figures \ref{fig:Variance20RelEntPlot} 
and \ref{fig:Variance20ALVRPlot}, the joining of the two smaller circles to the
group of the large circle is indicated by the second and third local maxima and minima,
respectively.

\subsection{Medium Noise Experiment ($\sigma = 0.03$)}

\begin{figure}[H]
\includegraphics[width=\textwidth]{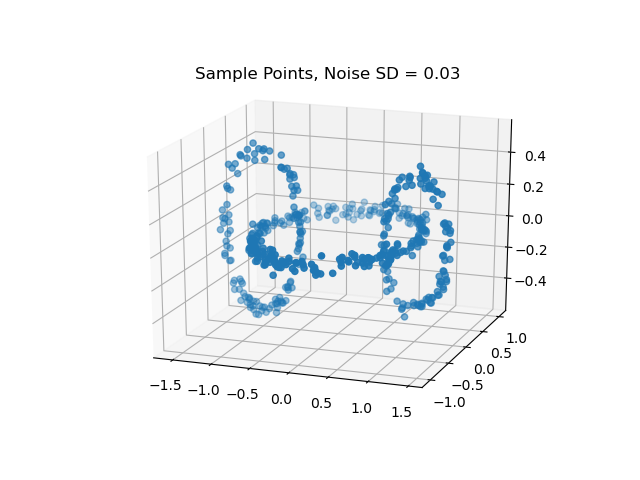}
\caption{500 points sampled from three circles with Gaussian noise ($\sigma =
0.03$)}
\label{fig:Variance30Points}
\end{figure}

\begin{figure}[H]
\includegraphics[width=\textwidth]{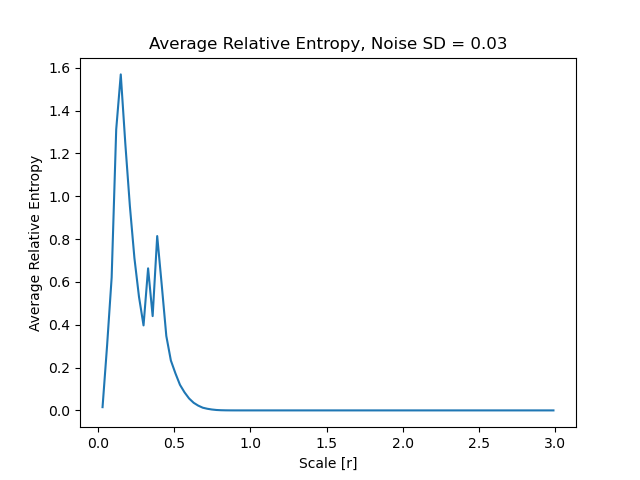}
\caption{Average Relative Entropy vs. Scale, Medium Noise Experiment. As
	before, the
local maxima appear where the smaller circles join to form a larger cluster.}
\label{fig:Variance30RelEntPlot}
\end{figure}

\begin{figure}[H]
\includegraphics[width=\textwidth]{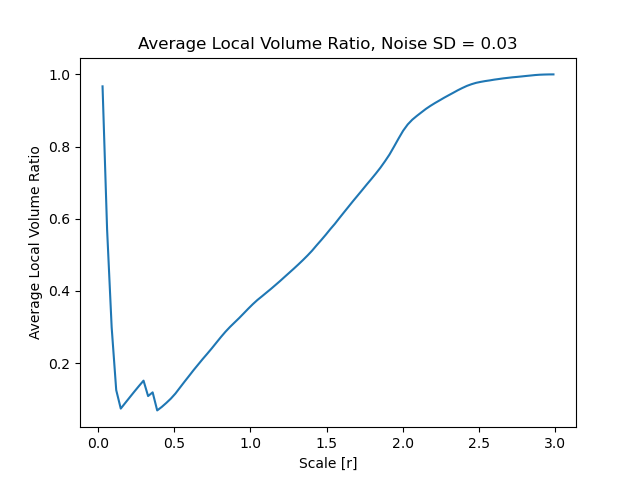}
\caption{Average Local Volume Ratio vs. Scale. Note that local minima appear
where the smaller circles join with the larger circle, and the global minimum
is the third local minimum.}
\label{fig:Variance30ALVRPlot}
\end{figure}

\begin{figure}[H]
\includegraphics[width=\textwidth]{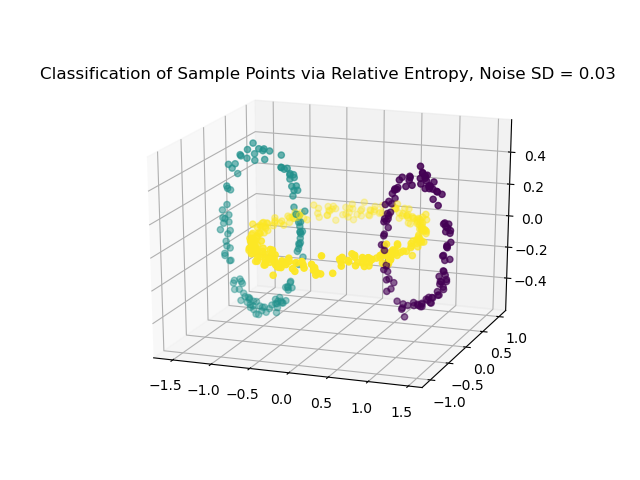}
\caption{The classification of the points using the Average Relative Entropy
method, illustrated by color. The Average Local
Volume Method assigned all of the points to a single class in this experiment.}

\label{fig:Variance30Clusters}
\end{figure}

In this experiment, the Average Relative Entropy Method successfully recovered
the circles, and the Average Local Volume Ratio Method returned a single
cluster of all three circles. Figure
\ref{fig:Variance30Points} illustrates the data set, Figure
\ref{fig:Variance30RelEntPlot} shows the Average Relative Entropy at each
scale, Figure \ref{fig:Variance30ALVRPlot} gives the plot of the Average Local
Volume Ratio, and Figure \ref{fig:Variance30Clusters} shows the
classification of the points. The image of $\Psi$ is roughly identical
to the previous experiment, so we do not repeat the plot here.  Note that, in contrast to the previous experiment, the
global minimum of the Average Local Volume Ratio occurs at the third local
maximum.

\subsection{High Noise Experiment ($\sigma = 0.045$)}

\begin{figure}[H]
\includegraphics[width=\textwidth]{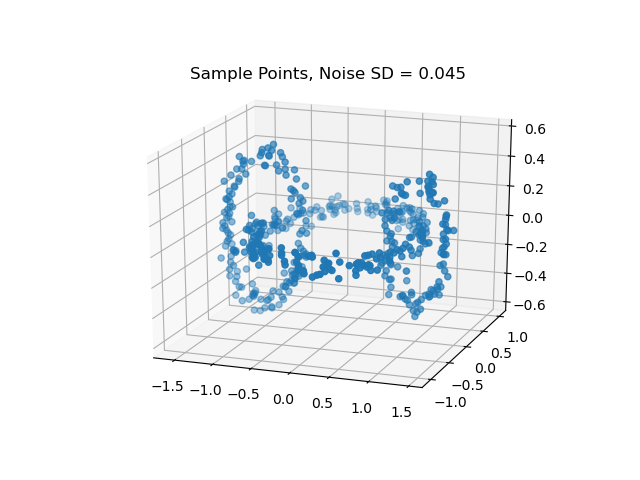}
\caption{500 points sampled from three circles with Gaussian noise ($\sigma = 0.045$)}

\label{fig:Variance45Points}
\end{figure}

\begin{figure}[H]
\includegraphics[width=\textwidth]{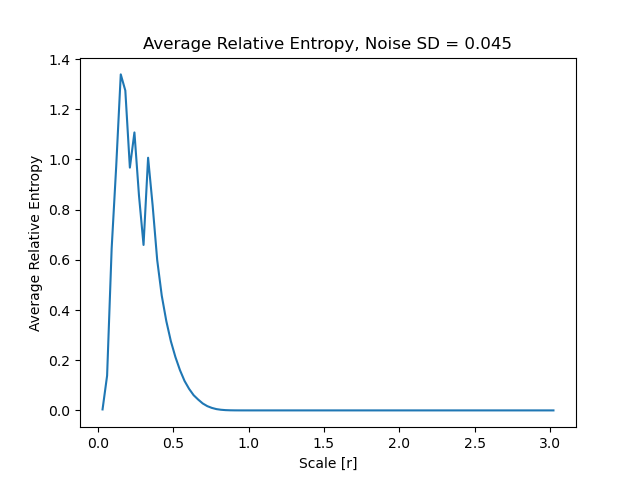}
\caption{Average Relative Entropy vs. Scale, Medium Noise Experiment. As
	before, the
local maxima appear where the smaller circles join to form a larger cluster.}
\label{fig:Variance45RelEntPlot}
\end{figure}

\begin{figure}[H]
\includegraphics[width=\textwidth]{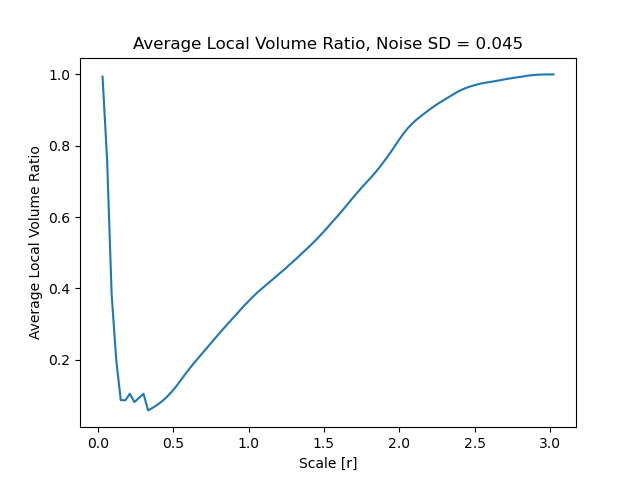}
\caption{Average Local Volume Ratio vs. Scale. As before, local minima appear
where the smaller circles join with the larger circle, and as in the previous
experiment, the global minimum is the third local minimum.}
\label{fig:Variance45ALVRPlot}
\end{figure}

\begin{figure}[H]
\includegraphics[width=\textwidth]{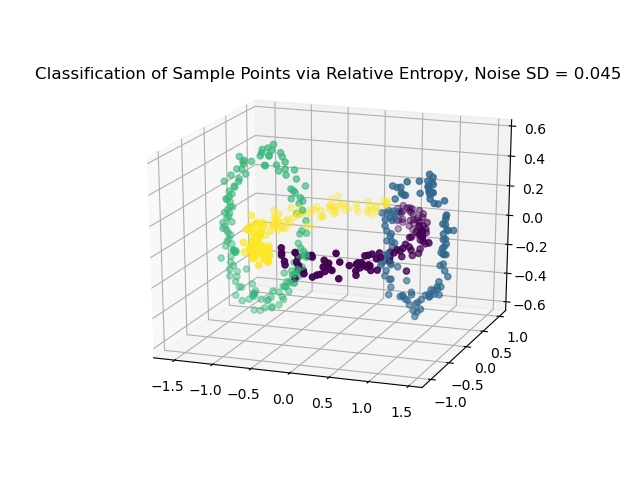}
\caption{The classification of the points using the Average Relative Entropy
method, illustrated by color. The Average Local
Volume Method assigned all of the points to a single class in this experiment.}

\label{fig:Variance45Clusters}
\end{figure}

In this experiment, we see the manner in which the Average Relative Entropy
Method begins to break down. While it successfully identified the two smaller
circles, the large, central circle is split into two clusters. While not ideal,
the circles could nonetheless be reconstructed from this clustering. As in the medium
noise experiment, the Average Local Volume Ratio Method returned a single
cluster of all three circles. Figure
\ref{fig:Variance45Points} illustrates the data set, Figure
\ref{fig:Variance45RelEntPlot} shows the Average Relative Entropy at each
scale, Figure \ref{fig:Variance45ALVRPlot} gives the plot of the Average Local
Volume Ratio, and Figure \ref{fig:Variance45Clusters} shows the
classification of the points by color. The image of $\Psi$ is roughly identical
to the previous experiment, so we do not repeat the plot. As in the medium
noise experiment, the global minimum of the Average Local Volume Ratio is found
at the third local minimum, i.e. after all of the circles have been joined to
the same cluster.

\section{Discussion and Future Work}

We have presented two novel data clustering algorithms for data sampled from a
uniform distribution on a disconnected metric space $X$, possibly corrupted by
Gaussian noise. Both algorithms work by identifying a scale $\hat{r}$ with
which to build a graph on the points, after which they identify the connected
components of the graph using the associated graph Laplacian. Unlike other
commonly used clustering algorithms, this technique is completely data driven,
and does not require any additional parameters. In particular, the
algorithms output the
number of clusters as well as the clustering, unlike the popular $k$-means
algorithm, which requires the number of clusters as input. Of the two
algorithms presented, the Average Relative Entropy Method outperformed the
Average Local Volume Ratio Method in terms of robustness to noise. 

We remark, however, that the success of these particular algorithms, is highly dependent on the
assumption of uniformity of the underlying, non-noisy distribution. This,
unfortunately, prevents the current form of these algorithms from producing a
correct clustering when applied to many datasets, and the
adaptation of these techniques to non-uniform distributions is the subject of
ongoing research. We also note that the Average Relative Entropy Method may be seen as a
variation of Diffusion Maps \cite{Coifman_Lafon_2006} and Laplacian Eigenmaps
\cite{Belkin_Niyogi_2003}, where the choice of free parameter is done in an
automatic, data-driven fashion, which allows for clustering to be achieved
directly from the eigenvectors of the operators, instead of after a
dimension-reduction step. While our method depends on the kernel function
having compact support, which is not the case in either Diffusion Maps or
Laplacian Eigenmaps, and in this paper we have concentrated on the clustering
problem, it would be interesting to extend the applicability of these
model selection methods to dimension-reduction problems as well.

\subsection*{Acknowledgements}We would like to thank Robert Adler, Jacob
Abernathy, and Bertrand Michel for useful discussions and
comments on an earlier, preliminary version of this work. We would additionally like to
thank the organizers of the CIMAT Differential Geometry Seminar, the organizers
of the AUS-ICMS Meeting, the XXIst Oporto Meeting on Topology, Geometry, and Physics, and the
Toposys Network for the opportunity to present our preliminary results in their
conferences and seminars.

\bibliography{/home/antonio/Bib/all}

\end{document}